\documentclass[11pt, a4paper, logo, twocolumn, copyright]{googledeepmind}

\pdftrailerid{redacted}

\makeatletter
\renewcommand\bibentry[1]{\nocite{#1}{\frenchspacing\@nameuse{BR@r@#1\@extra@b@citeb}}}
\makeatother

\usepackage{kantlipsum, lipsum}
\usepackage{dsfont}
\usepackage{dm-colors}

\usepackage{algorithm}
\usepackage{algpseudocode}
\usepackage{adjustbox}
\usepackage{multirow}
\usepackage{nicefrac}
\usepackage{cleveref}
\usepackage{soul}

\crefformat{section}{\S#2#1#3} 
\crefformat{subsection}{\S#2#1#3}
\crefformat{subsubsection}{\S#2#1#3}

\newcolumntype{R}[2]{%
    >{\adjustbox{angle=#1,lap=\width-(#2)}\bgroup}%
    l%
    <{\egroup}%
}

\usepackage[authoryear, sort&compress, round]{natbib} 
\usepackage{comment}

\newcommand{\todo}[2]{\textcolor{red}{\textbf{TODO: #1}: #2}}

\newcommand{\aar}[1]{\textcolor{teal}{{\bf AAR:}#1}}

\usepackage{pifont}
\newcommand{\cmark}{\ding{51}}%
\newcommand{\xmark}{\ding{55}}%

\newcommand{\Model}{DiPaCo}

\usepackage{subcaption}
\usepackage{url}

\graphicspath{{figures/}}

\title{DiPaCo: Distributed Path Composition
}

\correspondingauthor{douillard@google.com}


\reportnumber{} 

\renewcommand{\today}{2023-07-01} 

\author[*,1]{Arthur Douillard}
\author[*,1]{Qixuan Feng}
\author[*,1]{Andrei A. Rusu}
\author[1]{Adhiguna Kuncoro}
\author[1]{Yani Donchev}
\author[1]{Rachita Chhaparia}
\author[1]{Ionel Gog}
\author[$\diamondsuit$,1]{Marc'Aurelio Ranzato}
\author[$\diamondsuit$,1]{Jiajun Shen}
\author[$\diamondsuit$,1]{Arthur Szlam}

\affil[*]{Equal core contributions}
\affil[$\diamondsuit$]{Equal leading contributions}
\affil[1]{Google DeepMind}

\begin{abstract}
Progress in machine learning (ML) has been fueled by scaling neural network models.  This scaling has been enabled by ever more heroic feats of engineering, necessary for accommodating ML approaches that require high bandwidth communication between devices working in parallel. 
In this work, we propose a co-designed modular architecture and training approach for ML models, dubbed DIstributed PAth COmposition (\Model).   During training, \Model{}  distributes computation by paths through a set of shared modules.  Together with a Local-SGD inspired optimization (DiLoCo) that keeps modules in sync with drastically reduced communication, 
Our approach facilitates training across poorly connected and heterogeneous workers, with a design that ensures robustness to worker failures and preemptions. At inference time, only a single path needs to be executed for each input, without the need for any  model compression.  We 
consider this approach as a first prototype towards a new paradigm of large-scale learning, one that is less synchronous and more modular.

Our experiments on the widely used C4 benchmark show that, for the same amount of training steps but less wall-clock time, \Model{} exceeds the performance of a 1 billion-parameter dense transformer language model by choosing one of 256 possible paths, each with a size of 150 million parameters.

\end{abstract}

\begin{document}

\maketitle

\section{Introduction} \label{sec:intro}

Progress in machine learning and AI has been driven by spending more FLOPs on larger neural network models, trained on bigger data sets.  This scaling has been accomplished via data and model parallelism~\citep{NIPS2012_6aca9700} and pipelining~\citep{pipedream-2bw} to distribute computation, enabling the concurrent use of a large number of devices~\citep{openai2023gpt4, touvron2023llama, geminiteam2023gemini}.  Although model architectures \citep{lepikhin2021gshard, openai2023gpt4} have also been used to allow computational parallelism, and optimization procedures to prefer larger batches \citep{goyal2017accurate} (again allowing more data parallelism), the current training paradigm has not fundamentally changed model architecture or optimization procedure to facilitate distributed training. 
State of the art models are still essentially monoliths, and their optimization requires exchanges of parameters, gradients, and activations at every step of the learning process.

This approach incurs engineering and infrastructure challenges associated with provisioning and managing the large number of tightly interconnected devices required for
the lengthy training process.  The training process itself is often restarted for each new model release, essentially discarding much of the computation for training the last model.  Moreover, training monoliths incurs human-organizational challenges, as it is difficult to localize the effects to the final model of changes to any step in the process (beyond data preparation). 
In particular, it is difficult to leverage the potential of the greater ML community, rather than a single organization~\citep{raffel_ml_oss}. 
Because of these challenges, it may become more difficult to continue to scale with the current approach.

\begin{figure}[t!]
\centering
    \includegraphics[width=1\linewidth,trim={0cm 0cm 0cm 0cm},clip]{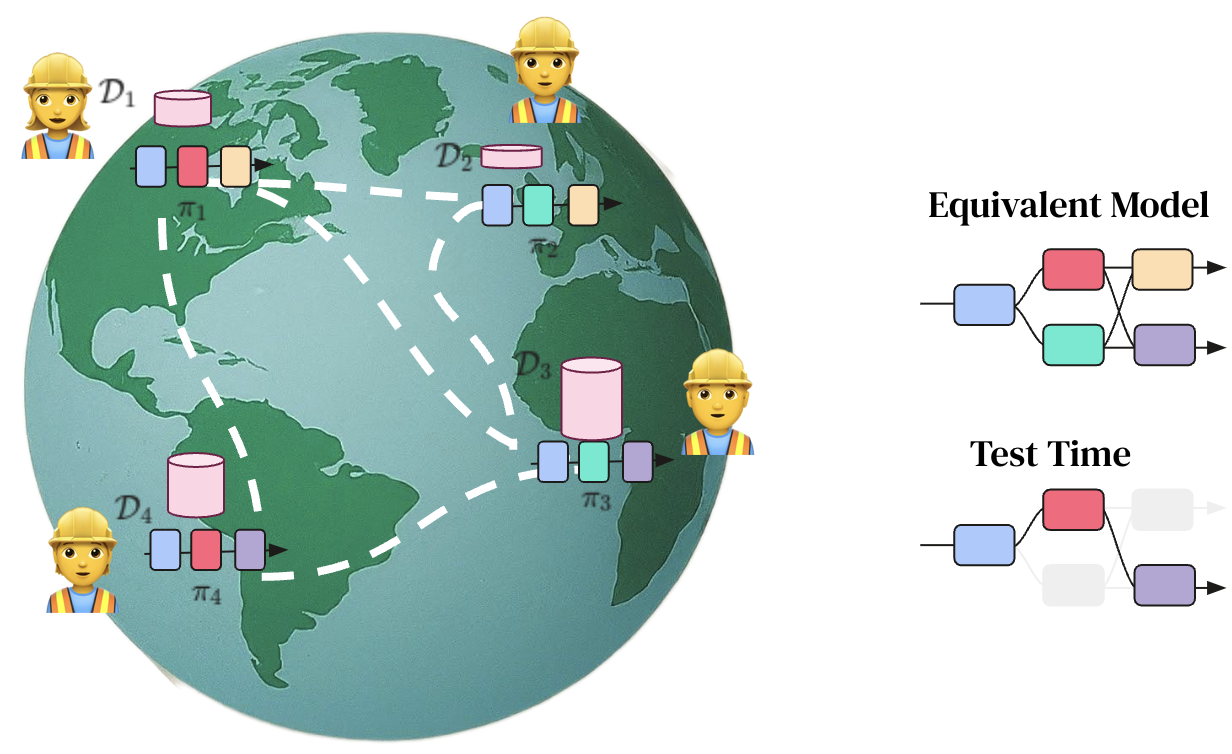}
    \caption{\textbf{Long-term Goal}: Ultimately, we envision a modular network where different components, \textit{paths} $\pi_i$, are optimized for different tasks, $\mathcal{D}_j$, each designed by different researchers. The paths, trained on any available hardware type, communicate infrequently across the world, exchanging useful information and enabling new forms of composition.}
\label{fig:dream}
\end{figure}

As in \citep{ryabinin2020towards, barham2022pathways, borzunov2022petals, raffel_ml_oss} and depicted in \autoref{fig:dream}, we envision an alternative approach where models are scalable both in terms of ability to ingest and train on large amounts of data and in terms of enabling collaboration with many contributors, and can be continuously updated and expanded, by virtue of modular design.  Modular ML may have many other benefits, see \cite{pfeiffer2023modular} for further discussion and references.

In this work, we take a step towards this scalable modular ML paradigm, 
proposing an architecture and training algorithm, DIstributed PAths COmposition (\Model).
\Model's architecture and optimization have been co-designed to reduce communication and enable better scaling. The high level idea is to \textit{distribute computation by path}; here, a ``path'' means a sequence of modules that define an input-output function. 
Paths are small relative to the entire model, thus requiring only a handful of tightly connected devices to train or evaluate. 
During both training and deployment, a query is routed to a replica of a path rather than a replica of the whole model; in other words, the \Model{} architecture is sparsely activated.   We give a detailed exposition of \Model{} in \autoref{sec:framework}, and the infrastructure we used to implement it in  \autoref{sec:infra}.

In \autoref{sec:experiments} we demonstrate the feasibility of \Model{} by training a language model on the C4 dataset~\citep{c4} with paths of size 150 million parameters, matching the performance in terms of validation perplexity of a 1.3 billion model, but with 45\% less wall clock training time.  While the dense 1.3B system required the use of all co-located devices, \Model{} uses 256 islands of compute, each of which is one-eighth the number of devices used to train the baseline.  Neither during train or evaluation time is it necessary to co-locate the full \Model{} model.

\section{Approach} \label{sec:framework}
In this section we give a detailed description of \Model{}.  We start by describing the assumptions and setting in which we will work.  
\subsection{Setting}
Our goal in this work is to demonstrate scalable ML models in the setting of many smaller islands of compute, as opposed to one tightly connected compute landmass.  Accordingly, we assume: \newline
1. Training compute (FLOPs) is relatively cheap\newline
2. Communication is relatively expensive\newline

These assumptions are {\it not} realistic in the current ML training paradigm.  
As discussed above, devices are usually co-located and run by a single organization, FLOPs are expensive, constrained by the purchase and installation of many accelerators, but communication between devices is relatively cheap because of the co-location.  

Nevertheless, in our view, 
these assumptions
may be realistic in the near future, if our compute requirements (or model sizes) grow beyond what can be reasonably co-located, and so communication costs become relatively more expensive.   
Conversely, progress in distributed training of ML models may allow simpler infrastructure build-outs, leading eventually to more available compute.  As it is, infrastructure is designed around the standard approach to training large monoliths; and ML models are architected to take advantage of the current infrastructure and training approaches.   This feedback loop may be leading the community to a spurious local minimum where compute is more constrained than it needs to be.

In addition (and closely related) to the two assumptions above, we also assume that both during train and evaluation, we cannot instantiate models as large as we would like to have on any single compute island.  

We will work in the context of language modeling (LM).
Based on the above assumptions, we will evaluate models by evaluation perplexity (PPL) against wall-clock time, keeping evaluation complexity in mind .  We choose this metric because it encourages exploring how to distribute model training, and more generally, points to in our opinion plausible new paradigms for ML.

\subsection{Overview of System}
The core idea of \Model{} is to train a sparsely-activated modular system where data and computation are distributed by the choice of path through the modules.  There are two key ideas to making this work:

\paragraph{Coarse Routing:}
Sparsely routed Mixture of Experts (MoE) have shown great results in language modeling \citep{lepikhin2021gshard}.  The standard approach in a transformer MoE LM is to make a routing decision at each token, based on the feature at each routed layer.  In contrast, in this work, during training we will route once per document and offline, as in ~\citep{gross17, gururangan2023scaling}.  

Routing once per document allows batching computation across all tokens of a sequence, without the need to swap modules in and out as a sequence is processed. This in turn allows parameters to be distributed across distant workers.
In addition, instead of learning the router along with the model, we will compute routing decisions
{\em offline}. This enables pre-sharding data by path, before the start of training. This is critical to distribute training across paths, as each worker can now  train a path by processing its own shard of data, using its own set of parameters. In \autoref{sec:coarse_routing}, we will describe in detail how we approach coarse routing.

\paragraph{DiLoCo to keep modules in sync}
Paths cannot be trained completely independently, because some modules might be shared across multiple paths.  To support module sharing across paths, we use DiLoCo~\citep{douillard2023diloco} for low communication data parallelism, see \autoref{sec:local_sgd}.  If there are $P$ paths (assigned to $P$ workers) that share module $i$, each corresponding worker performs SGD on its own shard of data, and every {\em few hundred} steps, workers average the difference of the parameters of module $i$ before and after the local SGD phase.  These averages are then used to update a global parameter vector which is then re-distributed across the workers to sync them, a procedure described in \autoref{sec:multilevel_moe}.

With these two choices, 
neither at training nor at test time does the entire network (collection of paths) need to be materialized together, as shown in  \autoref{fig:dream}.

\subsection{Notation} \label{sec:notation}
In this section we introduce the basic notation used throughout this work. 
We assume that we have a base model architecture with parameters $\theta$.  We take a partition of the parameter indices into $L$ subsets $B_l, l \in [1,\ldots,L]$. 
For each $l$, we choose a number $K_l$ that represents the number of possible choices for the parameters in $B_l$.  We will call a set of parameters associated to a $B_l$ a ``module'' or, as in ~\citep{jacobs91, jacobs94}, an ``expert''.  In this terminology,  $K_l$ is the number of distinct modules associated to $B_l$.

\begin{figure*}[ht!]
    \begin{minipage}{.49\textwidth}
    \includegraphics[width=0.9\linewidth,trim={6cm 10cm 9cm 8cm},clip]{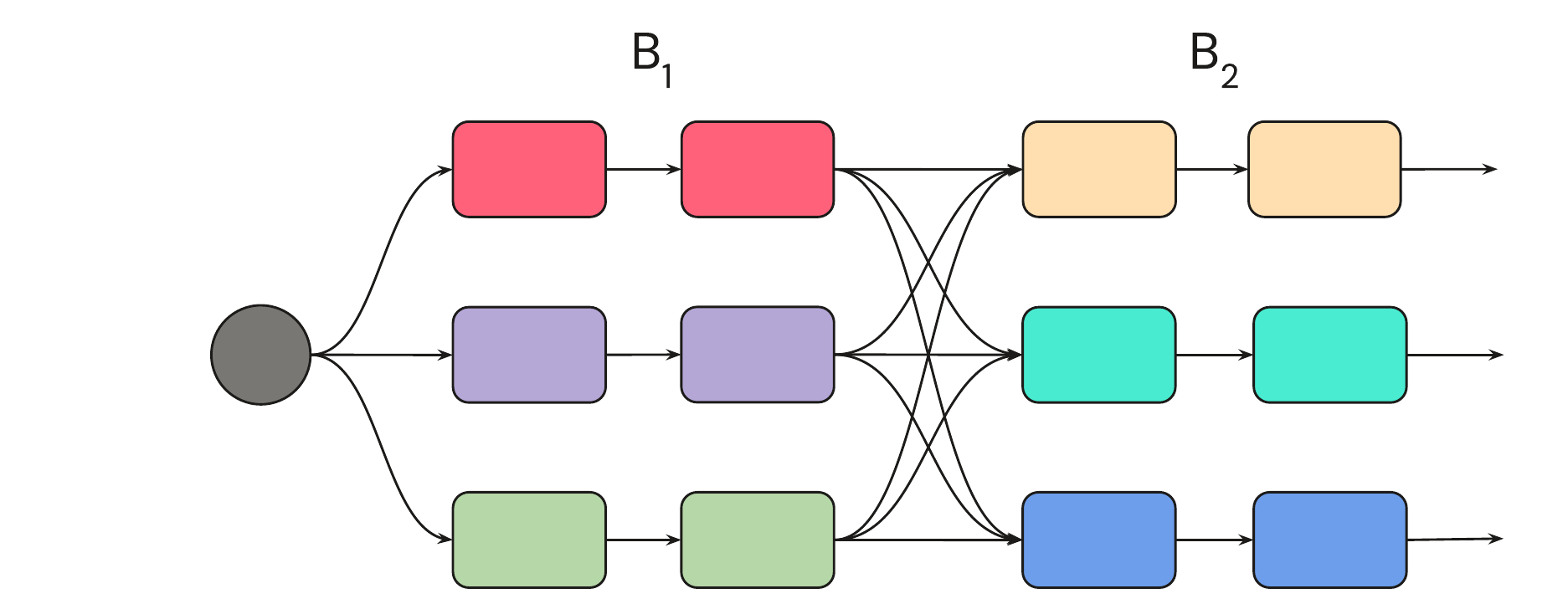}
    \end{minipage}
    \begin{minipage}{.49\textwidth}
    \includegraphics[width=0.9\linewidth,trim={6cm 10cm 9cm 8cm},clip]{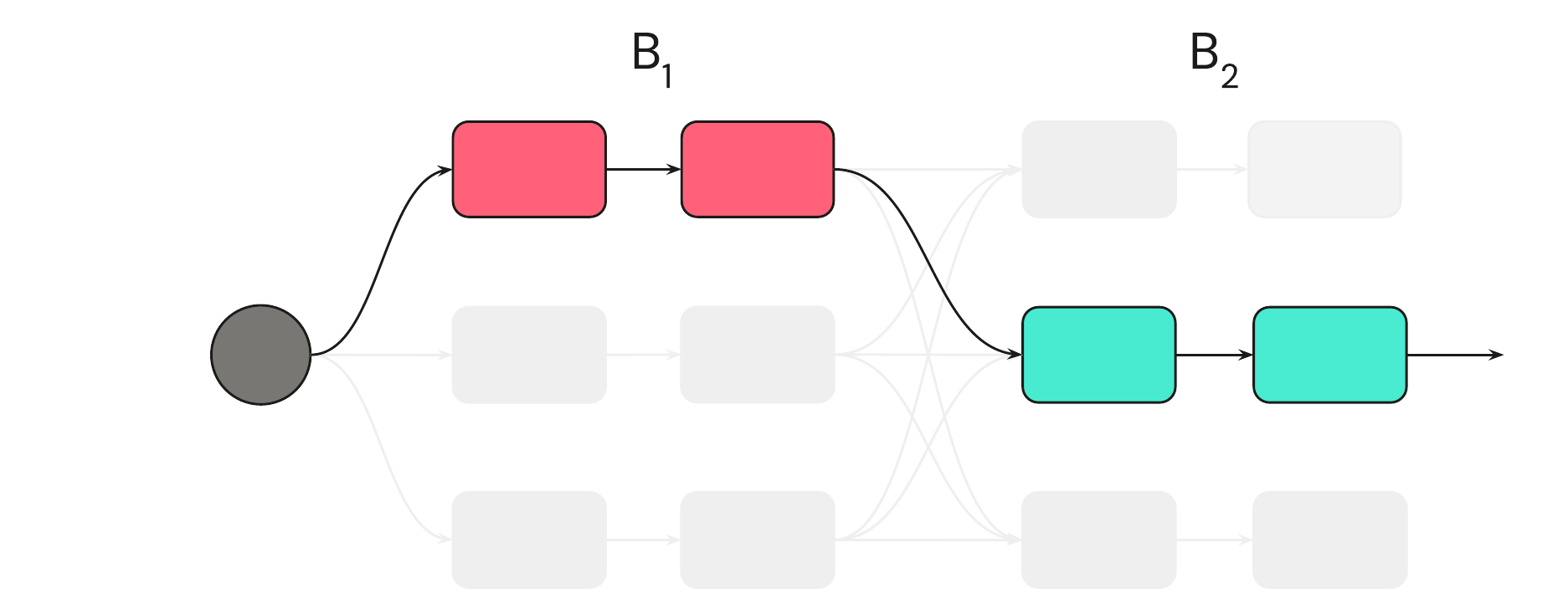}
    \end{minipage}
\caption{An illustration of the first example from Section \ref{sec:notation}.  A $4$ layer neural network, with block $B_1$ consisting of the first $2$ layers and $B_2$ consisting of the next $2$ layers.  Each block has $3$ choices of module (each with its own parameters), represented by different colors.  On the left, we show all of the $9$ possible paths. On the right, we show a single path.  }
\label{fig:framework}
\end{figure*}

For a simple example, consider a four-layer fully connected network $A$.  We might take the first two layers to be $B_1$ and the third and fourth layers to be $B_2$, so $L=2$.  If we choose $K_i=3$ for $i=1, 2$, then we have a $3$ modules in $B_1$, and $3$ modules in $B_2$,  for a total of $6$ modules.   Any choice of module from $B_1$ and module form $B_2$ defines a neural network; as in \cite{pathways}, we call each of these $9$ possible networks a ``path'', see \autoref{fig:framework}.

In this example, and most of the cases we will consider in the rest of this paper, the ``blocks'' $B_i$ determine contiguous sub-networks of $A$ that can act as an input-output mapping, but this is not necessary.  We could just as easily let $B_1$ be layers $1$ and $3$ and $B_2$ be layers $2$ and $4$; or $B_1$ be all the biases in the network, and $B_2$ be all the linear parameters.  Nevertheless, we will call $i$ the ``level'' of the module. 

In order to transform an input $x$ to an output $y$, we need to figure out which parameters will be used to operate on $x$.  The function that takes $x$ to a choice of parameters will be called the ``router'', and is denoted by $r$; thus $r(x)$ has the form $[j_1, ..., j_L]$, where each $j_l$ indexes one of the $K_l$ choices of parameters for $B_l$, and $r$ maps an input to a path $\pi = \pi_{j_1,\ldots,j_L}$.  We will also often collapse the sequence $j_1,...,j_L$ into a single number $j$, where $j \in \{1, \ldots, P=\Pi_{i=1}^L K_i\}$.  If the training dataset is denoted by $X$, the subset of data that is routed to path $j$ will be called the $j$-th ``shard'' $\mathcal{D}_j$.

When the router maps an $x$ to a distribution over $\{1, \ldots, P\}$ rather to a deterministic choice, we say that the router is ``soft''.  In this work, we will consider ``hard'' routers instead, where the choice is single valued and deterministic.  

Finally, it is useful to refer to module parameters. Assume that the $i$-th path goes through module $e$ at level $l$, with $e \in [1, K_l]$, $l \in [1, L]$.
Since during training paths are not fully synchronized, we denote the local copy of the parameters of module $(l,e)$ in path $i$ at iteration $t$ with $\theta(l, e)^t_i$. The collection of parameters used by the $i$-th path (across all levels) is denoted by $\theta^t_i$. When modules across paths are synchronized, the parameter values of module $(l,e)$ is the same across all paths, and therefore we denote the global copy of the parameters of module $(l,e)$ at iteration $t$ by omitting the path index, as in $\theta(l,e)^t$. Similarly, we denote local and global gradients of module $(l,e)$ with $\Delta(l, e)^t_i$ and $\Delta(l, e)^t$, respectively.

\subsection{Coarse Routing} \label{sec:coarse_routing}
In this section we describe how we route sequences.
The basic intuition is to use some context to decide which path is most suitable to process a given sequence. Since we focus on language modeling, we use as context the first $32$ tokens of each sequence. At training time, we then use the remaining tokens for learning the model parameters. At test time, we use the remaining tokens to compute perplexity on a held-out validation set. All methods, including dense baselines, will be evaluated in the same way, by calculating perplexity using all but the first $32$ tokens of each sequence, which was used to determine the routing decision of our models.

Notice that the outcome of routing is assigning each sequence to a certain path in the network. Collectively, all sequences that are assigned to a particular path form a shard of the original dataset. In this work, there is a one-to-one association between paths and corresponding shards; path $\pi_i$ is associated to shard $\mathcal{D}_i$.

Next, we describe how we use this prefix of $32$ tokens to route, as well as other variants of routing.

\begin{figure*}[ht!]
\centering
    \includegraphics[width=1\linewidth,trim={0.5cm 4.5cm 2cm 3cm},clip]{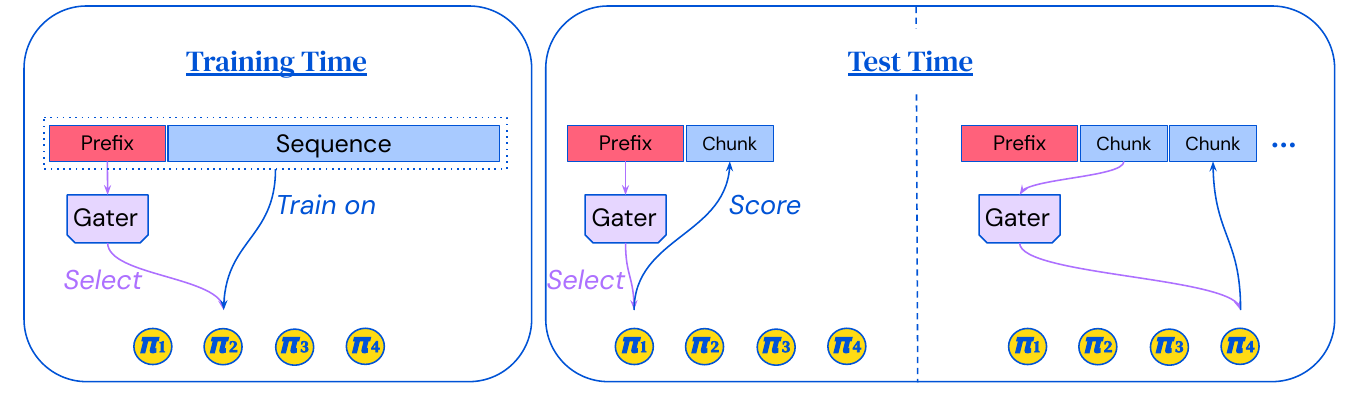} 
    \caption{\textbf{routing More Frequently at Test-Time}: At training time (left panel), the router selects the path $\pi_i$ using the prefix $z$. We train the chosen path on the whole sequence using the usual language modeling loss. At test time (right panel), the path selected by the router given the prefix is used to score the next chunk of tokens. Then, we re-use the router to choose the most likely path given the new chunk. This process repeats until the whole sequence has been scored.}
\label{fig:frequent_gating}
\end{figure*} 

\subsubsection{Generative routing} \label{sec:generative}

In generative routing, the decision about which shard to assign a sequence to is not informed by the task at hand, namely language modeling. Instead, the choice is based on minimizing feature reconstruction error. Given a representation $z$ of the first $32$ tokens of a sequence (context), we perform $k$-Means on the features $z$ of each sequence, and then we use the $k$-means assignment algorithm to shard the data into $k$ shards. If $\{c_1, \dots, c_k\}$ are the $k$ prototypes learned by $k$-means, the sequence with prefix $z$ is assigned to shard $\mathcal{D}_{r(z)}$ via: 
\begin{equation}
r(z) = \arg \min_{i\in[1,k]} ||z - c_i||^2. \label{eq:kmeans}
\end{equation}

\subsubsection{Discriminative routing} \label{sec:disc_gating}
Generative routing is agnostic of the language modeling task. In discriminative routing the sharding takes into account how well experts perform on each sequence. The training set is split into two parts. The first and largest part is used to train the paths on data sharded using a generative sharding appproach. The second and much smaller part is used to evaluate paths on each sequence. Let $s_i$ be the log-likelihood score of the path $\pi_i$ on the sequence associated with prefix $z$. The index of the path attaining the largest likelihood score, $i^* = \arg \max_{j} s_j$, is used as label for $z$. We then train a logistic regression classifier to directly predict $i^*$ from the features $z$; this trained classifier is used as $r$.
Finally, we use such classifier to re-shard the entire training and validation datasets. 

This process is an approximation to Expectation Maximization~\citep{em}, a coordinate descent method alternating between updating the parameters of the model (paths), and latent variables (path assignments). It is an approximation because, for simplicity, we use as label the top scoring path as opposed to the full posterior distribution over path scores. Of course, this re-assignment process can be repeated for as many times as desired.

We invite those readers  interested in diving deeper in the routing mechanism to read \autoref{app:routing} of the Appendix for further details.

\subsubsection{Routing More Frequently at Test Time} \label{sec:freq_gating_test}
Routing only once per sequence is critical to distribute training across paths. However, at test time we can afford routing more frequently. Like \autoref{fig:frequent_gating} shows, we can divide a sequence at test time into chunks of $W > 1$ consecutive tokens, e.g., $W=128$. We can then feed a router with the $i$-th chunk to predict the best path to use at the  $(i+1)$-th chunk, using a process similar to the one described in \autoref{sec:disc_gating}. The cost of switching paths is small, since the router is run infrequently and in scoring (as opposed to auto-regressive generation) mode. Moreover, only text needs to be communicated to the router and next selected path.

\subsubsection{Overlapping Shards}
\label{sec:overlapping_shards}
Above, each sequence was associated to one and only one shard; in other words, we pre-sharded the dataset into a set of $k$ disjoint partitions.

When the number of shards $k$ is large, the mixture might overfit, particularly when paths share few parameters. 
To combat this issue we might assign each sequence not just to one shard but to its $n$ ``closest'' shards. 
This makes shards overlap with each other, smoothing out the boundaries among themselves, which may improve generalization when the number of shards is relatively large.  This can be done {\it independently} at train time and at evaluation time; that is, one might overlap shards only during train, at eval, or both, or neither.  At train time, overlapping the shards limits the specialization of each path, and increases the storage necessary for each shard, but otherwise does not have a computational cost, as each path is still trained independently on its shard.  However, overlapping the shards at evaluation means that the model has to be forwarded through multiple paths, which increases the evaluation computational cost. 

In this work, when we overlap the shards in training, we use the top-2 choices, while we never overlap the shards at evaluation.  We leave to future work other approaches to  overlap shards.

\subsection{DiLoCo: Review} \label{sec:local_sgd}
In this section, we summarize DiLoCo~\citep{douillard2023diloco} as it lays the foundation of our distributed optimization algorithm, which will be described in the next section. Together with coarse routing of \autoref{sec:coarse_routing}, DiLoCo is a critical component of \Model. 

DiLoCo optimizes a dense model across $k$ workers. First, the original dataset is sharded into $k$ shards, and each shard is associated to a worker. Second, all workers start from the same copy of the model parameters (typically a pretrained model), and perform $H$ (inner) optimization steps over their own shard independently. Third, each worker sends to a central CPU server the difference in  parameter space between their new parameters and their initial ones. We refer to these differences as \textit{outer gradients}. The central server averages these outer gradients into a single update vector. Finally, the global parameter vector is updated using an (outer) optimizer, and the new global parameter vector is re-dispatched to all workers for another phase of local training. The process repeats for as many rounds as desired.

In language modeling applications using transformers, the inner and outer optimizers that have been shown to be most effective are respectively AdamW~\citep{kingma2014adam} and Nesterov momentum~\citep{sutskever2013nesterov}. Empirically in prior work, $k$ has been set up to 64 and $H$ up to thousand. Note that other alternatives, such as FedOpt \citep{reddi2021adaptive}, are compatible with this framework.

\subsection{\Model{}} \label{sec:multilevel_moe}

\begin{figure*}[ht!]
\centering
    \includegraphics[width=0.9\linewidth,trim={.3cm 6cm 0cm 8cm},clip]{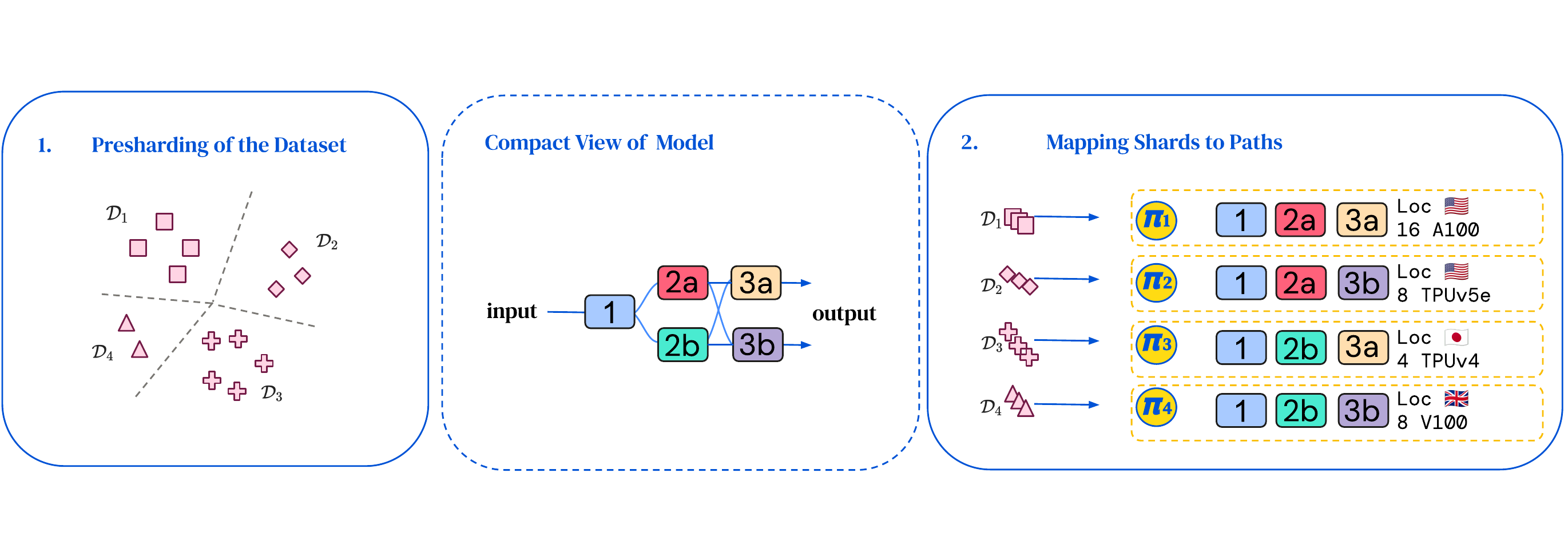}
    \caption{\textbf{\Model}: (left)  The dataset is pre-sharded into $k$ shards, $\mathcal{D}_i$ (here $k=4$). (middle)  Compact view of a $2\times2$ \Model, which is never instantiated. In this toy illustration, there are three levels. Level 2 and 3 have a mixture with two modules each. Level 1 has a single module shared by all paths. (right) 
    We associate each shard $\mathcal{D}_i$ to a \textit{path} $\pi_i,\, \forall i \in [1, 4]$. In this toy illustration, a path is the composition of three neural network blocks. The color refers to the id of a module. The figure shows the modular network unrolled across the four paths. These are trained by using DiLoCo which requires communicating only every few hundred steps. In this example, module 2a (in \textcolor{dmred500}{red}) is shared by paths $\pi_1$ and $\pi_2$. Workers associated to paths might use different  hardware types (different kind of GPUs or TPUs) and might be placed in far away geographic areas.}
\label{fig:moeml}
\end{figure*} 

In \autoref{sec:coarse_routing}, we described methods to pre-shard data, and in \autoref{sec:local_sgd} we described how a model can be trained across several workers that communicated infrequently.
This section combines these two approaches towards the efficient training of a composable mixture, a network whose levels are replaced by a mixture of experts. We dub such architecture and training algorithm \Model.  In our experiments below, a level is composed of several consecutive transformer blocks. 

In the toy illustration of \autoref{fig:moeml} there are three levels, $B_1$, $B_2$, and $B_3$. There is only one module (equivalently, one set of parameters) in $B_1$, and it is shared across all paths. There are two modules in $B_2$ (meaning that there will be two distinct parameter vectors for the module at that level); the first is shared by  paths $\pi_1$ and $\pi_2$, and the second by paths $\pi_3$ and $\pi_4$. Similarly, at the third level $\pi_1$ and $\pi_3$ share parameters, and $\pi_2$ and $\pi_4$ share parameters.  
 
The resulting $2 \times 2$ \Model{} has $4$ paths in total (as shown on the middle panel of \autoref{fig:moeml}). However, the full model need never be fully instantiated, neither during training nor testing.  Only paths are realized (as shown on the right panel of \autoref{fig:moeml}), and these are  trained in parallel with infrequent communication.  

At training time DiLoCo is applied at the level of modules, as opposed to the entire network; see lines 11-16 of \autoref{alg:local_sgd_moe}. The outer optimizer updates independently the parameters of each module after receiving the corresponding outer gradients from workers processing paths passing through that particular module. For instance and with reference to \autoref{fig:moeml}, the outer gradient of module 3b is calculated by averaging the gradients of  module 3b of the worker processing $\pi_2$ and $\pi_4$. Such  outer gradient is then used to update the parameters of module 3b.
A worker does not need to be powerful enough (in terms of memory and compute) to host the entire model, but just a single  path. Like in DiLoCo work~\citep{douillard2023diloco}, the inner optimizer is AdamW and the outer optimizer is Nesterov momentum. 
Similarly, at test time, the paths are instantiated, and served independently, with text routed to each path via a router.

\begin{algorithm*}
\caption{\Model{} (see notation in  \cref{sec:notation})} \label{alg:local_sgd_moe}
\begin{algorithmic}[1]
\Require Num. levels $L$, num. experts per level $K_l$, paths $\pi_i$ with $i \in {1,\ldots,P}$ and $P=\prod_{i=1}^L K_i$. Let $P_{l,e}$ be the number of paths going through module $e$ at level $l$.
\Require Pre-sharded training set into $\{\mathcal{D}_1, \ldots,  \mathcal{D}_P \}$ (see \cref{sec:coarse_routing}).
\Require Parameters of pretrained model used for initialization: $\bar{\theta}$. $\theta_i^0 = \bar{\theta}, i \in {1, \ldots, P}.$
\Require Optimizers $\texttt{InnerOpt}$ and $\texttt{OuterOpt}$
\For{\texttt{outer step $t = 1 \ldots T$}}
  \State (Optional, in this work done once during training) discriminatively re-shard data (see \cref{sec:disc_gating})
  \For{\texttt{worker $i = 1 \ldots P$}}
  \State $\theta_i^t = \theta_i^{t-1}$
    \For{\texttt{inner step $n = 1 \ldots \tau$ }}  
        \State $x \sim \mathcal{D}_i$
        \rlap{\hspace{5.55cm}\smash{$\left.\begin{array}{@{}c@{}}\\{}\\{}\\{}\\{}\\{}\\{}\end{array}\color{dmblue300}\right\}%
          \color{dmblue300}\begin{tabular}{l}\textbf{Inner Optimization}: parallel per path\end{tabular}$}}
        \State $\mathcal{L} \gets f(x, \theta_i^t)$
        \State $\theta_i^t \gets \texttt{InnerOpt}(\theta_i^t, \nabla_\mathcal{L})$ 
    \EndFor
  \EndFor
  
  \For{\texttt{level $l = 1 \ldots L$}}
      \For{\texttt{expert $e = 1 \ldots K_l$}}
        \State $\Delta(l,e)^t \gets \frac{1}{P_{l,e}} \sum_{i=1}^{P_{l,e}} (\theta(l,e)^{t-1} - \theta(l,e)_i^t)$ 
        \rlap{\hspace{0.45cm}\smash{$\left.\begin{array}{@{}c@{}}\\{}\\{}\\{}\\{}\\{}\end{array}\color{dmblue300}\right\}%
          \color{dmblue300}\begin{tabular}{l}\textbf{Outer Optimization}: parallel per module\end{tabular}$}}
        \State $\theta(l,e)^t \gets \texttt{OuterOpt}(\theta(l,e)^{t-1}, \Delta(l,e)^t)$  
  \EndFor
\EndFor
\EndFor
\end{algorithmic}
\end{algorithm*}

\subsubsection{Increasing Capacity}
\label{sec:capacity}
\begin{figure}[th]
\centering
    \includegraphics[width=1.0\linewidth,trim={7cm 3cm 9cm 5cm},clip]{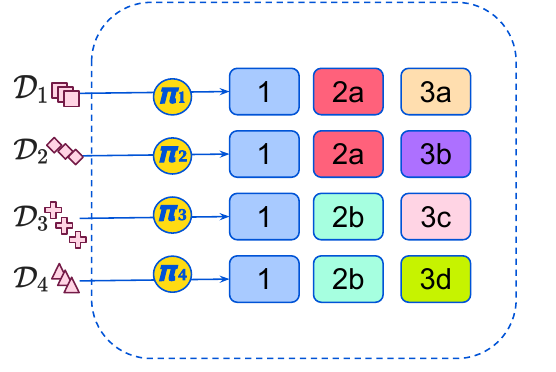} 
    \caption{\Model{} with more capacity: In this example, level 3 modules are path specific, i.e., modules at that level are not shared by paths.}
\label{fig:moem_more_capacity}
\end{figure} 
The more paths go through a module, the more opportunities for transfer learning across paths, but also the more constrained learning is and the less capacity the overall mixture has. \Model{} gives a lot of flexibility in how to set this trade-off. Whether a level is shared by all paths or by a certain subset only does not change the overall framework. The only difference is in the choice of which subset of paths is used when computing the average in line $13$ of \autoref{alg:local_sgd_moe}.

One extreme choice is to allocate path-specific modules, as shown in \autoref{fig:moem_more_capacity}. In this case, the third level has a mixture with as many modules as there are paths. This is a particularly easy way to increase parameter count in \Model.\footnote{While there is no gradient averaging anymore for these path-specific modules, we still apply the outer optimization of \autoref{alg:local_sgd_moe} because empirically it improves convergence over the default optimizer.}

Clearly, the proposed structure of \Model{} is merely a first step.  We consider it especially exciting to co-design the optimization methods and the architecture.  In the next section, we discuss some of our initial approaches to adapting optimization methods for our setting.

\subsubsection{Scaling the modular architecture}\label{sec:extreme_scaling}

\Model{} is made of multiple levels, each with several expert modules. \textit{e.g} a $16\times16$ has two levels, which contains 16 modules on each level, producing in total 256 paths. It is easy to scale the architecture size by increasing the number of levels and modules per levels, \textit{e.g} a $32\times 32 \times 32$ has $32{,}768$ paths. It's difficult to find enough devices for that number of paths, therefore only a subset of the paths can be trained at any moment in time. Potentially, at each start of inner optimization, we could sample a different subset of the paths. Doing so would allow us to scale \Model{} to arbitrary large size.

\subsubsection{Flat Mixture of Experts} \label{sec:flat_moe}
The extreme form of capacity increase as in \autoref{sec:capacity} would be to have each path be a completely independent network.  We will refer to this model as a \textit{flat} mixture of experts (flat MoE). In a flat MoE paths do not share \textit{any} parameter. In the language of \autoref{sec:notation}, there is only one level, and $B=B_1$ is the entire network, and $K=K_1$. 

Flat MoE is a strong baseline for the  compositional version of \Model.  In our experiments, we see that it can outperform architectures with shared modules when the number of shards is small compared to the size of the total amount of data.

 With generative routing using $k$-means, this approach is essentially ~\citep{gross17, gururangan2023scaling}; in our experiments below we will use discriminative routing as in the compositional models, as this consistently outperforms generative routing.

\subsection{Advanced Optimization Techniques}\label{sec:optim_moeml}

\paragraph{Outer Gradient Norm Rescaling:}

In \Model, each module can belong to a different number of paths.  For instance, in a $16\times 16$ \Model, 16 paths can contribute to a single module, while there could be a level that has path-specific modules which is not shared by any other. 

Consequently, the average outer gradients, $\Delta(l,e)^{t}$ in \autoref{alg:local_sgd_moe}, have significant different norms for different modules. Using the intuition, and also empirical observation, that averaging across a larger number of paths is akin to using a larger batch size, we have rescaled the outer gradient norm by the square root of the number of paths going through a module. 

\paragraph{Loss reweighing:} 
In general, shards might have different amounts of data. If training of \Model{} is distributed by using one worker per shard/path, then according to the uniform average of line $13$ in \autoref{alg:local_sgd_moe} smaller shards are over-sampled.
To compute an unbiased estimate of the gradient, we therefore weigh the outer gradients proportionally to the shard size:
\begin{equation}
\Delta^{(t)}_{l,e} \gets \sum \alpha_{l, e} (\theta^{(t-1)}_{l,e} - {\theta_{l,e}^{(t)}}), 
\end{equation}
with:
\begin{equation}
\alpha_{l, e} = \frac{|\mathcal{D}_{l, e}|}{\sum |\mathcal{D}_{l', e'}|}.
\end{equation}

\paragraph{Early Stopping:}
By setting aside a small subset of training examples in each shard, we can perform path specific early stopping. In this case, for each path we selected the parameters that yield the lowest loss on the corresponding shard validation set.  We found that early stopping improves generalization on small shards, e.g., when the number of shards is large.

\section{Infrastructure} \label{sec:infra}

\begin{figure*}[ht!]
\centering
    \includegraphics[width=0.8\linewidth,trim={0cm 0cm 0cm 0cm},clip]{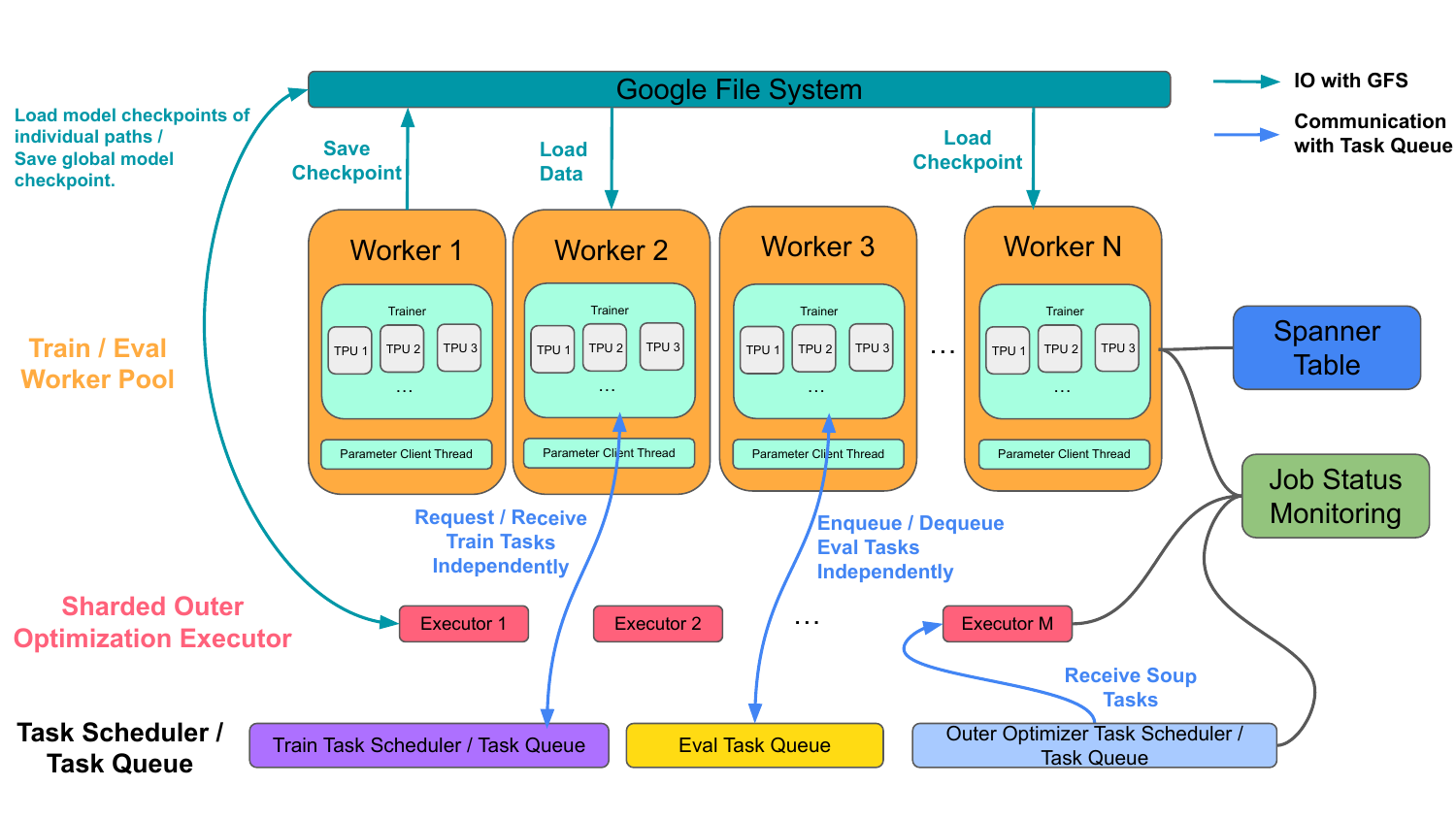} 
    \caption{\textbf{Infrastructure}. Training workers (in \textcolor{orange}{orange}) responsible for performing the inner optimization fetch training task from the train task queue (in \textcolor{violet}{purple}). Once the training checkpoints are saved to GFS, their paths and metadata (e.g. training step, path ID, phase ID) are written to a Spanner table (in \textcolor{blue}{blue}) allowing easy look up of checkpoint path given metadata. The evaluation workers and the sharded outer optimization executors (in \textcolor{red}{red}) load training checkpoints once they become ready, as signaled by their appearance in the Spanner table. Finally, a monitoring worker (in \textcolor{green}{green}) periodically checks the health of workers and reboots them if necessary.}
\label{fig:infra}
\end{figure*} 

We now describe the infrastructure we designed to implement the framework detailed earlier. The infrastructure should achieve the following objectives: 1) reliably train the modular architecture via infrequent synchronization; 2) ensure easy scalability for scenarios involving training across multiple pods or data centers; and 3) provide continuous fault tolerance, allowing every component of the system to quickly recover from failed hosts and preemptions without halting training. 

We illustrate the different components in \autoref{fig:infra}. The general training workflow is the following:
\begin{enumerate}
    \item At the beginning of each training phase,
    a list of training tasks, each including pairs of path and shard ids, is added to the training task queue maintained by the task scheduler server (in \textcolor{violet}{purple}).
    
    \item When a training worker in the worker pool (in \textcolor{orange}{orange}) becomes available, it fetches the next training task from the train task scheduler and performs inner optimization (L5-9 of \autoref{alg:local_sgd_moe}) on accelerators. Once the inner optimization is finished, the checkpoint is saved to the Google's distributed file system \citep{ghemawat2003gfs} and the training worker becomes available for the next training task. The path to the checkpoint, along with the metadata of the checkpoint (e.g., path ID, outer step ID, etc.), is recorded in a database table (shown in \textcolor{blue}{blue}). This enables other components to query the checkpoint file path for a given path.
    
    \item When a checkpoint is saved by a training worker (indicated in \textcolor{teal}{teal}), evaluation tasks for that checkpoint are added to the eval task queue (highlighted in \textcolor{yellow}{yellow}). The eval task queue is then consumed by the evaluation workers in the worker pool to carry out evaluation tasks.
    
    \item The outer optimizer task scheduler (indicated in \textcolor{cyan}{light blue}) distributes outer optimization tasks to sharded outer optimization executors (highlighted in \textcolor{red}{red}), each of which is responsible for the outer optimization of a shard of modules (e.g., a single module or a collection of modules). Each executor loads training checkpoints containing the corresponding modules as soon as they appear in the Spanner database table~\citep{corbett2012spanner} and performs the parameter averaging and outer optimization (L13-14). The checkpoints of the updated module parameters are saved to the distributed file system, and their paths with metadata are written to the database table.
    
    \item After completion of all the training tasks, the current training phase concludes, and the subsequent training phase commences and training tasks in the new phase can be started as soon as their corresponding modules finish the outer optimization step. See \ref{sec:sharded_soup_exec} for more detail.

    \item Throughout training, a job status monitor (in \textcolor{green}{green}) periodically checks the health of workers and task queue servers, and restarts them if they become unresponsive.
\end{enumerate}

All the components are designed to be robust to failures and pre-emptions. We will discuss more details of each infrastructure component in the following subsections.

\subsection{Worker Pool} \label{sec:worker_pool}

We employ a producer-consumer design pattern to distribute the training tasks. Within each phase, training tasks are added to the task queue, each of which involves training a path for a specific number of steps from a given checkpoint. We maintain a pool of workers, where each worker retrieves training tasks from the task queue iteratively. Each training task is completely independent of other tasks, requiring no synchronization or communication among the workers.

In the event of worker failure or preemption, the fault-tolerant task queue server would return the task from the unavailable worker back to the task queue before reassigning it to another available worker. The task queue server also periodically checkpoints the current task queue, making it possible to recover from server failures or preemptions.

The key advantage of this design is that, thanks to the complete independence among workers, the system can continue making progress even if some workers become unavailable, as long as the worker pool is not empty. It is worth noting that the worker pool can contain {\em heterogeneous} types of devices across {\em different regions}, making it easy to scale up further. Additionally, the worker pool allows for elastic resource utilization by auto-scaling pool size according to resource availability.

\subsection{Task Queue System} \label{sec:mh_task_queue}

In practice, the aforementioned task queue system is designed to manage and distribute asynchronous tasks among the workers in the worker pool using remote procedure calls. It consists of three components: the task queue scheduler, the task queue server, and the task queue client. Training tasks are published by the task queue scheduler and sent to the task queue server, while each worker instantiates a task queue client to request tasks from the task queue.

In the multi-host training scenario, where multi-host parallelism is achieved by Single Program, Multiple Data (SPMD), initializing a task queue client on each host would result in each host pulling a different task from the task queue. We solve this issue by implementing a task queue client that synchronizes its results across all hosts. This is achieved by creating a real task queue client only on the first JAX host. All other hosts get the response from the first host by performing a blocking all-gather operation.

We also need to synchronize task queue write operation outside of the main Python thread in some places. For instance, evaluation tasks for a given checkpoint can only be enqueued after checkpointing on all hosts has finished. For this use case, we implemented a barrier, which blocks until each program running on their host have made a call with the same unique key.

\subsection{Outer Optimization Efficiency} \label{sec:sharded_soup_exec}

\begin{figure}[t!]
\centering
    \includegraphics[width=1.0\linewidth,trim={2cm 4cm 11cm 6cm},clip]{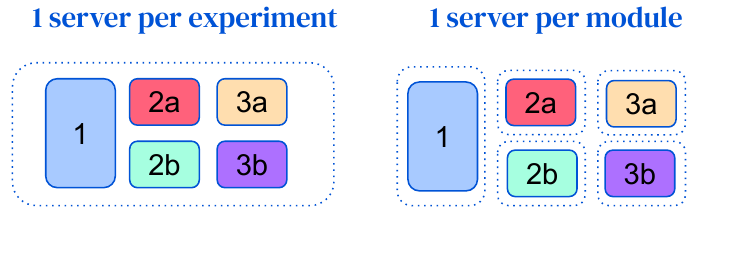} 
    \caption{\textbf{Sharded Outer Optimization Executor}: An example of outer optimization executors sharded by module, which significantly reduces the processing time and memory requirement.}
\label{fig:sharded_soup_exec}
\end{figure} 

Na\"ive implementation of outer optimization executor introduces significant overhead as we scale up the number of paths across the world. The following optimizations allow us to scale to hundreds of paths with average time per phase for outer update under 2 minutes (lines $11$-$16$ of \autoref{alg:local_sgd_moe}).

\paragraph{Online Parameter Gradient Averaging:} As the training tasks within a phase may not finish at the same time, it is possible to reduce  the parameter averaging time by loading and accumulating a training checkpoint to the current partial sum as soon as one becomes available. It is also possible to extend to an asynchronous update that doesn't require to wait for all paths before doing an outer update \citep{liu2024asyncdiloco}.

\paragraph{Sharded Outer Optimization Executor:} As our model architecture is modular, averaging meta-gradients of different modules can be performed independently. By distributing the parameter averaging across multiple servers, as illustrated in \autoref{fig:sharded_soup_exec}, we reduce the memory requirement of each averaging task, as well as the overhead of checkpoint loading and averaging. It also allows to reduce the training time as a training task can be started as soon as its corresponding modules have finished their outer update. As a consequence, the overall model is never materialized in a single location but always split across several servers.

\paragraph{Asynchronous Checkpoints Gathering:} Each outer optimization executor loads as many checkpoints as the number of paths sharing its associated module. It is possible to load checkpoint from anywhere on earth from the distributed file system \citep{ghemawat2003gfs}. However, if the checkpoint, located on the server where training took place, is at a certain distance from the outer optimization executor, there is going to be a significant delay. In this case we launch in the background an Effingo process \citep{google2023effingo} to bring the checkpoint to a closer location.

\paragraph{Miscellaneous Improvements:} In addition to the previous speed improvements, we also a) cache the parameters of the outer optimizer \texttt{OuterOpt}, b) reuse the just-in-time compiled optimizer, and c) load multiple paths in parallel via a multi-threading queue.

\subsection{Backup Pool} \label{sec:backup_pool}

By default we create as many training workers (see in \textcolor{orange}{orange} in \autoref{fig:infra}) $\mathcal{W}_i$ as paths/shards of data, $\forall i \in [1, P]$. However as we scale the number of paths, the hardware requirement might become prohibitive. For instance, with 256 paths and 16 \texttt{A100} GPUs per worker, we would need $4\,096$ GPUs. Therefore in practice, the number of workers is often less than the number of paths, and instead, we do multiple \textit{rounds} of training within an outer iteration step until all paths have been trained (L$3$-$10$ of \autoref{alg:local_sgd_moe}).

To minimize the number of rounds, and therefore maximize speed, we create a backup pool of training workers. Namely, we spawn as many workers as there are paths across multiple accelerator types using a low-tier priority. As soon as an accelerator is available, it is snatched and used for an inner optimization phase. Although device with low-tier priority are preempted frequently, we can benefit from the backup pool since each training task only takes roughly a couple of minutes (the inner optimization L$5$-$9$ in \autoref{alg:local_sgd_moe}).

Note that this backup pool is particularly useful when used with the path sampling idea mentioned in \autoref{sec:extreme_scaling}.
\section{Experiments} \label{sec:experiments}

We consider a language modeling task on the C4 dataset, derived from Common Crawl~\citep{c4}, tokenized with a SentencePiece tokenizer \citep{kudo2018sentencepiece} with a vocabulary size of $32{,}000$. We report perplexity on the validation set against number of weight update steps used at training time, which is a close proxy for wall-clock time if all computations are done on the same accelerator type. The total number of weight update steps is set to $88{,}000$. All paths on \Model{} have size $150$ million parameters, using a transformer with $12$ blocks, $896$ dimensional hidden states and $16$ heads.

We build modular networks of varying number of paths, and varying levels of module reuse.  We compare to the performance of a dense transformer language model of size $1.3$B, which has $24$ blocks, $2048$ dimensional hidden states and with the same number of heads; and a dense model that is the size of one \Model{} path. These models are also trained for $88{,}000$ weight update steps.

We warn the reader that this comparison is {\it not} standard in the literature, as weight updates for \Model{} see more tokens and use more FLOPs when the number of paths is larger.    
Nevertheless, while our modular networks is made of hundreds of paths, they are all trained in parallel, and thus our training wallclock time is 45\% less than the $1.3$B dense counterpart, and roughly equivalent to the $150$M parameter single-path counterpart.

In our experiments we have searched over relatively few hyper-parameters: mainly learning rate and value of Nesterov momentum. We use a sequence length of $1{,}024$ tokens and a batch size of $512$. At eval time, we consider sequences of $2{,}048$ tokens. We use cosine learning rate scheduling with peak value of $0.0004$. We list all the hyperparameters in the appendix (\autoref{sec:supp_details}).

All instances of \Model{} use one phase of discriminative routing (see \ref{sec:disc_gating} and \ref{sec:disc_routing_details}).  The $16\times 16 = 256$ path \Model{} uses top-2 overlapping shards (\ref{sec:overlapping_shards}) at training time (this does not make training slower in wallclock or FLOPs, although it does increase the size of the shards).

\subsection{Comparison with 1B dense model}
\begin{figure}
\centering
    \includegraphics[width=1\linewidth,trim={0cm 0cm 0cm 0cm},clip]{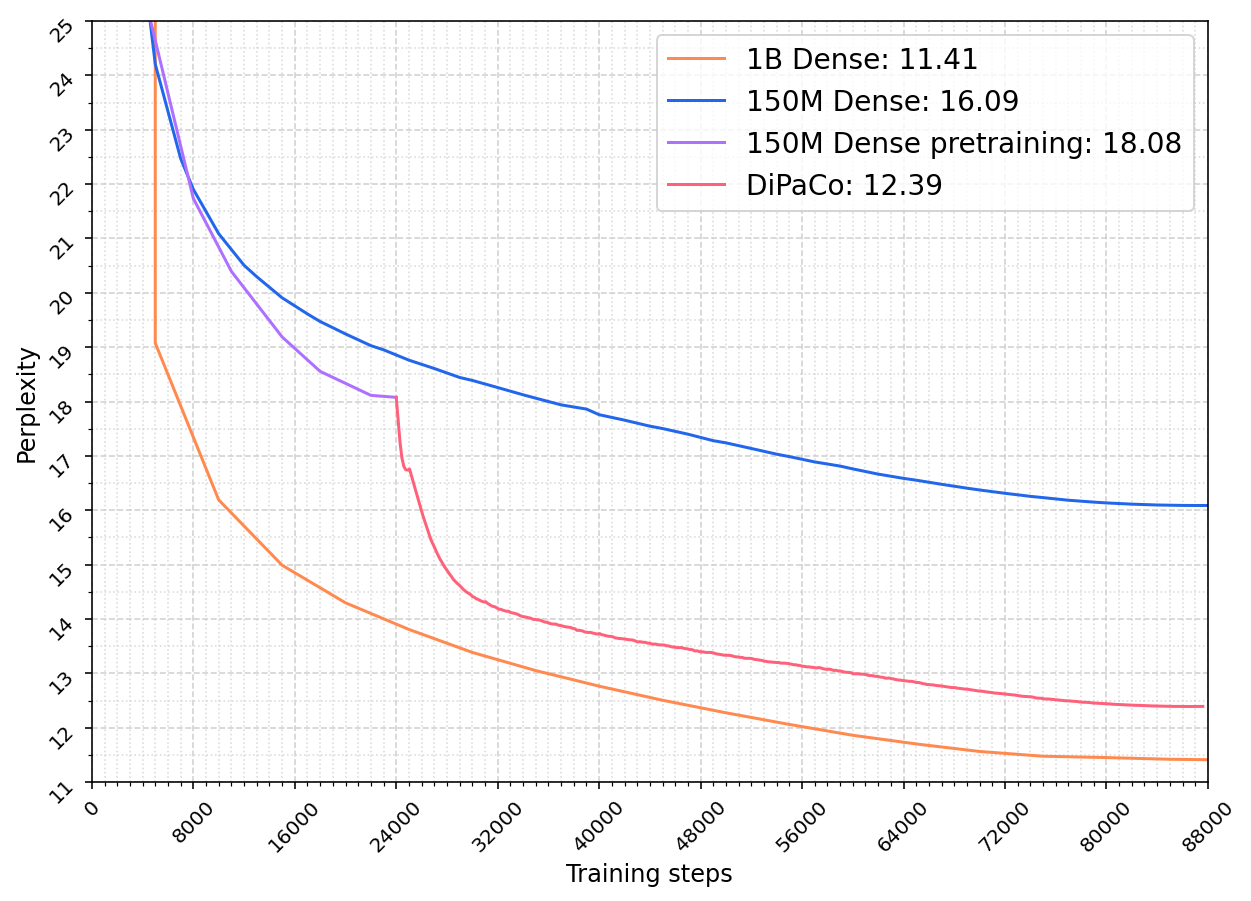} 
    \caption{\textbf{Convergence curves of \Model{} vs Dense Baseline}: We first pretrain a $150$M parameters model for 24k training steps (\textcolor{dmpurple500}{purple}). We then finetune a 16x16 \Model{} $P=256$ (\textcolor{dmred500}{red}). The remaining gap to the dense 1.3B parameters baseline (\textcolor{dmorange500}{orange}) is reached by gating more frequently 
     at test time.}
\label{fig:curves}
\end{figure} 
\autoref{fig:curves} compares the convergence curve of the 1.3B dense language model with the $16 \times 16$ \Model. 
\Model{} with $256$ paths of size $150$M partially sharing parameters, and one routing decision per document, is nearly sufficient to match the performance of a dense model of size $1.3$B. 
The remaining performance gap with the dense $1.3$B model can be closed with early stopping and routing every 64 tokens at test time, see \autoref{tab:frequent_gating_moeml}. 

We again warn the reader that the $x$ axis in \autoref{fig:curves} is not FLOP equivalent for the different models.  For the $150$M baseline model and for \Model{}, the $x$ axis roughly corresponds to equivalent wall-clock (and the $1.3$B model is slower w.r.t. wall-clock for the same number of updates).  On the other hand, \Model{} uses many more FLOPs and sees more tokens per training step than the dense baselines.  However, it is not trivial to parallelize training of the dense models to the scale of \Model{}, see the comparisons in \autoref{sec:parameter_sharing_exps} and in particular \autoref{fig:diloco_vs_dipaco}, and in \citep{douillard2023diloco}.

\subsection{Number of paths and parameters}
\label{sec:num_paths}
\begin{figure}
\centering
    \includegraphics[width=1\linewidth,trim={0cm 0cm 0cm 0cm},clip]{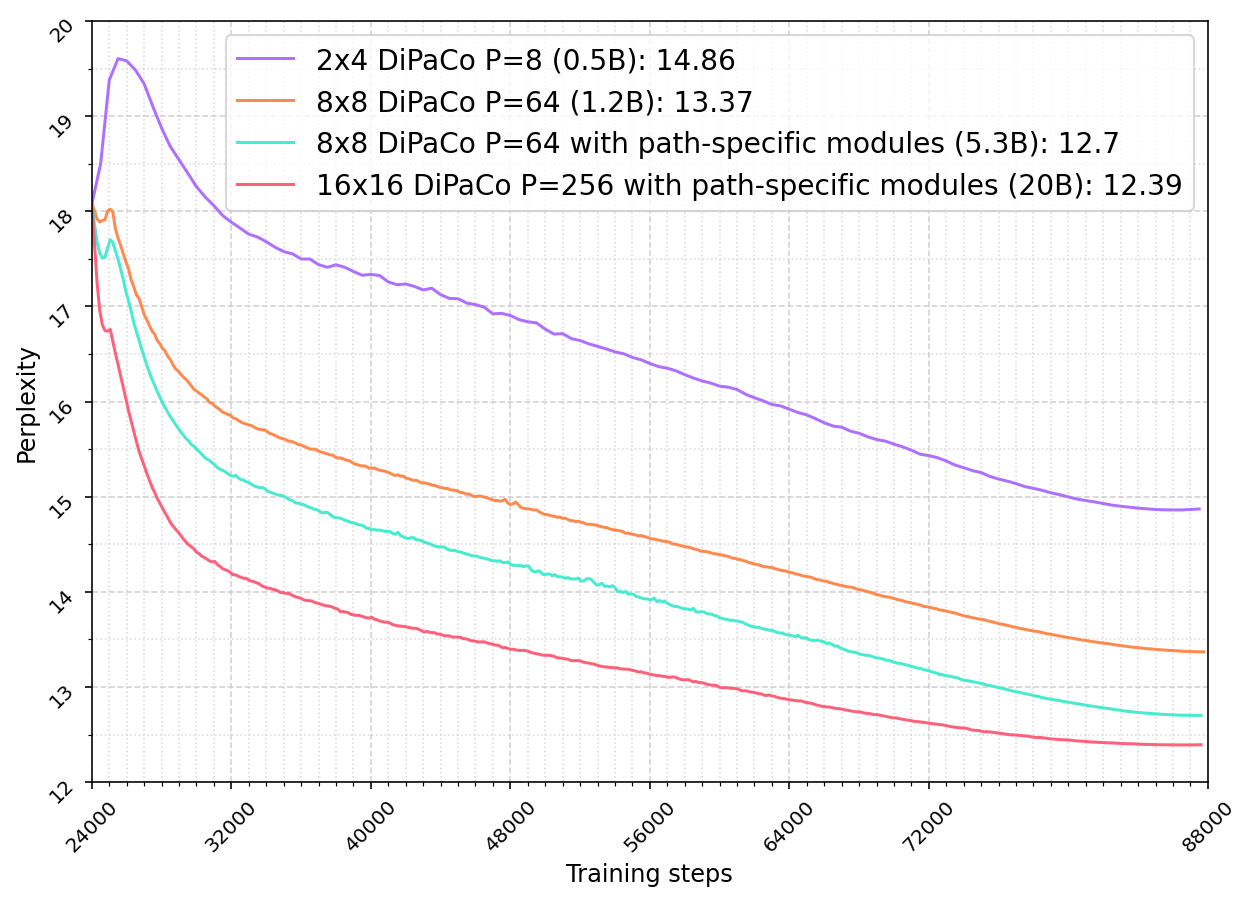} 
    \caption{\textbf{Scaling number of paths in \Model}: We report validation perplexity of \Model{} with different numbers of paths and parameters. We increase capacity by varying both the number of paths (from \textcolor{dmpurple500}{8} to \textcolor{dmred500}{256}) and by using path-specific modules. The size of a path (roughly equivalent to serving cost during deployment) is $150$M parameters for all models in this figure.}
\label{fig:scaling_moeml}
\end{figure} 
\autoref{fig:scaling_moeml} shows that generalization improves as more paths and more overall parameters are added to the mixture. Since the overhead of DiLoCo is minimal, the x axis could be replaced with wall-clock time. Of course, $16 \times 16$ \Model{} requires $32$ times more compute than $2 \times 4$ \Model{}, but these additional paths can be trained in parallel with very little communication. In this experiment the inner optimization consists of $\tau=150$ steps. For all experiments with \textit{path-specific modules}, the transformer blocks $0$, $5$, $6$ $11$, and the embedding matrix are not communicated across paths.

\subsection{Parameter sharing between paths}
\label{sec:parameter_sharing_exps}
\begin{table*}
\centering
\begin{tabular}{@{}l|cccc@{}}
\toprule
Model & Time & Compute and Data & Total Parameters & Validation PPL \\
\midrule
Baseline           & $1\times$ & $1\times$  & $150$M & 16.23  \\
\midrule
DiLoCo $P=8$       & $1\times$ & $8\times$  & $150$M & 15.02  \\
DiLoCo $P=64$      & $1\times$ & $64\times$ & $150$M & 14.96 \\
\midrule
Flat MoE $P=8$       & $1\times$ & $8\times$  & $1.2$B   & 14.62  \\
Flat MoE $P=64$      & $1\times$ & $64\times$ &  $9.6$B & 12.76 \\
\midrule
DiPaCo $2\times 4$ & $1\times$ & $8\times$  & $0.5$B   & 14.86  \\
DiPaCo $8\times 8$ & $1\times$ & $64\times$ & $1.2$B  & 13.37 \\
DiPaCo $8\times 8 + $ Path Specific Modules & $1\times$ & $64\times$ & $5.3$B  & \bf{12.70} \\
\midrule
\midrule
Baseline, $8\times$ steps   & $8\times$ & $8\times$ & $150$M & 14.72  \\
\bottomrule
\end{tabular}
\caption{\textbf{\Model{} vs. Flat MoE vs. DiLoCo}: We compare DiLoCo and Flat MoE (\ref{sec:flat_moe}) vs. \Model{}.  We also compare against the baseline trained for the same FLOPs as the distributed models.
In DiLoCo all paths are collapsed together at each outer optimization step. In Flat MoE, all paths are independent.  In \Model, only a subset of the parameters are ``collapsed'' using information from a subset of paths.  Because of the distributed framework, \Model{} is trained in the same wallclock time as DiLoCo, which is roughly the wallclock of the baseline.}
\label{fig:diloco_vs_dipaco}
\end{table*}

At one end of the spectrum, as described in \autoref{sec:flat_moe}, we use paths that share no parameters, and can be trained fully independently. Continuing the trend from \autoref{sec:num_paths}, where there is monotone improvement in validation performance as parameters are added, we see improvements by unsharing all parameters for small number of paths.  Using 64 unshared paths we can approach a $16\times 16$ \Model{} that has no unsharing.

However, this model starts over-fitting once the number of shards (and corresponding experts) is further increased as shown in \autoref{tab:flat_moe_overfit}.  We are able to recover some performance by overlapping shards, but still we find that we cannot naively scale the number of fully independent paths.  We consider approaches for more sophisticated overlapping of shards and path branching to be a promising direction for future work.

At the other end of the spectrum, paths can share all parameters, effectively collapsing back into a single path and recovering the performance of the original DiLoCo~\citep{douillard2023diloco}.  
We can see that the base model trained with DiLoCo, or base model overtrained for many steps, although receiving the same number of FLOPs and tokens as the larger \Model{} models, do not have the capacity to make use of these extra FLOPs.  Of course a properly sized baseline would do much better; and indeed the 1.3B model above already has better perplexity.  However, \Model{} consists of paths that each require only $150$M size island of compute both in train and testing, and its distributed training algorithm is built into the architecture, so that only paths of $150$M parameters ever need to be materialized on any training or deployment worker. We compare all these variants in \autoref{fig:diloco_vs_dipaco}.

\subsection{Routing frequency during evaluation}
We demonstrate the results of routing more frequently at test time in \autoref{tab:frequent_gating_moeml}. We observe a significant improvement of $0.74$ perplexity points going from routing once per sequence of $1{,}024$ tokens to every $128$ tokens. Further increasing the granularity of the re-routing consistently improve results, although more marginally so.
This shows that at training time we can route at the sequence level to enable pre-sharding for more efficient learning, and then at evaluation time we can gain back performance by routing more frequently. By routing tokens potentially to a different path every 64 tokens, \Model{} reaches $11.38$ perplexity, matching the performance of a $1$B dense model ($11.41$). Since at any time  \Model{} uses a single path of size $150$M parameters, it matches the performance of the $1$B model using $6$ times less parameters, and therefore, requiring less compute at inference time.

\begin{table}
\centering
\begin{tabular}{@{}l|cc@{}}
\toprule
\# fully independent paths & Validation PPL \\
\midrule
$P=8$ &  14.6  \\
$P=16$ & 13.9 \\
$P=256$ & 14.2 \\
\bottomrule
\end{tabular}
\caption{\textbf{Flat MoE (independent paths) overfits as number of paths increase}: Validation perplexity of 
\Model{} with no shared modules, as described in \autoref{sec:flat_moe} after $10$K steps of training.  With overlapping shards as described in \autoref{sec:overlapping_shards} and early stopping, we can improve the $P=256$ result to $13.6$, but still the models overfit at $64$K steps with $256$ paths.  In contrast, there is no overfitting at $256$ paths with overlapping shards with a $16\times 16$ \Model{}.}
\label{tab:flat_moe_overfit}
\end{table}

\begin{table}
\centering
\begin{tabular}{@{}cc|r@{}}
\toprule
Early Stopping & Route Every & Perplexity \\
\midrule
\xmark & \multirow{2}{*}{Once per sequence}  & 12.39 \\
\cmark &  & 12.22 \\
\midrule 
\multirow{4}{*}{\cmark} & 128 & 11.48 \\
                        & \textbf{64}  & \textbf{11.38} \\
                        & 32  & 11.31 \\
                        & 16  & 11.26 \\
\bottomrule
\end{tabular}
\caption{\textbf{Frequent Routing at Eval-Time}: Validation perplexity of a $16\times 16$ \Model{} with $P=256$. Despite using paths of size $150$M ($>6\times$ fewer parameters), we match the performance of a dense $1$B model ($11.41$) by potentially re-routing to a different path every $64$ tokens, although the model may choose to route to the exact same path that it selected previously.}
\label{tab:frequent_gating_moeml}
\end{table}

\subsection{Parameter synchronization with DiLoCo vs. true gradients}
Finally, we ablate the effect of using the partially synchronous optimization  DiLoCo to understand whether we lose performance by communicating less frequently at training time. In the setting where all devices are actually collocated and there is no constraint in terms of communication, we can train the \Model{} model in a fully synchronous manner without using DiLoCo: At every step, each path computes gradients on \textbf{its own batch of data} from its own data shard; Gradients across all paths are then exchanged and aggregated module by module; Finally, the model performs one step of AdamW update with the aggregated gradient. To our surprise, we did not observe a significant performance gap between the fully synchronous training setting and the partially synchronous training setting. More specifically, \Model{}  trained with DiLoCo slightly outperforms their fully-synchronously-trained version by $0.3$ and $0.6$ perplexity points when using a $2\times 2$ and $4\times 4$ architecture, respectively. At $8\times 8$ \Model{} trained fully synchronously reaches better perplexity by only $0.1$ perplexity despite communicating hundred of times more. This suggests that DiLoCo is an effective distributed optimization algorithm for \Model.

\section{Related Work} \label{sec:related_work}

As mentioned in the introduction, this work shares the same motivation and intuitions expressed in Pathways~\citep{pathways}. Unlike the Pathways framework~\citep{barham2022pathways} which supports training of general modular multimodal multitask asynchronous systems, we propose a particular instantiation of a modular system that supports such kind of distributed training. 
Our approach also shares motivations and intuitions with \cite{ryabinin2020towards, borzunov2022petals}.  The key difference in this work is that each worker trains a {\it path} through modules, rather than a module.

\subsection{Modularity}
There is a large body of literature on modularity, although this is not always framed in such terms.
At one end of the spectrum there are approaches that use a very large number of very tiny modules.
For instance, RETRO~\citep{retro} is one such approach, whereby there is one module per token n-gram of a large retrieval dataset, and the modules are merely vectors of biases.
At the other end of the spectrum there are approaches that use very few (e.g., two) but very large modules. For instance, Flamingo~\citep{flamingo} is a vision language model composed by two main modules, a pre-trained visual encoder and a pre-trained language model. Similarly, \cite{dalmia2023legonn} proposed another approach where modules are pre-trained language models and  speech recognition encoders. The authors  showed  how such modules could be plugged in different ways to enable 0-shot learning of entirely new tasks, such as producing a speech recognizer that translates into another language.

These lines of work are representative of the vast majority of the literature on modularity.
There is however also some work where the number of modules and their size is intermediate.
For instance, \cite{senthil19} showed how softly gated modules could be used in visual question answering to handle compositional tasks. In multi-task cross-lingual transfer, \cite{pfeiffer2020madx} used instead a network where modules are very small adapters, and the vast majority of parameters are shared across paths. 
We refer the reader to the survey by~\cite{pfeiffer2023modular} for additional pointers, and focus next on modularity via mixture models.

\subsection{Mixture of Experts}
Together with ensembling and boosting, mixture of experts (MoE) are an early approach to modularity in the machine learning  literature~\citep{jacobs91, jacobs94}. Unlike ensembling, mixture models could be more efficient if an input is processed only by a few experts. Unlike boosting, training of mixture models is efficient because it can be parallelized, while in boosting the process is inherently sequential.

While early work considered the problem of hierarchical gating of flat mixtures~\citep{jacobs94}, mixtures at multiple levels appeared only more recently. For instance, \cite{eigen14} proposed a two-level mixture where paths were softly mixed and trained jointly via stochastic gradient descent. In that paper, the authors recognized the difficulty of learning non-degenerate routing functions and discussed some workarounds. The issue of how to learn a good routing function has been a major topic of research in the field since then. 

In their seminal work, \cite{shazeer2017} proposed a very large mixture of experts LSTM model for sequence modeling tasks. Most works MoE for sequence-modeling works that followed~\citep{lepikhin2021gshard, switchtransformer, artetxe2021efficient, clark22} have used a recipe whereby FFN layers of transformers are replaced by mixtures.  These MoEs operate at the token level, and training operates similarly to dense model training: all paths are synchronized at every gradient descent step.
Token-MoE are currently state of the art with respect to training FLOP efficiency, but require even more co-located accelerators than the equivalent-activated dense model for training.  

In contrast, \citep{gururangan2023scaling} trains experts independently using a document level router; this approach had been used in computer vision by \citet{gross17}, and it also appeared in the federated learning literature~\citep{reisser2021federated}.  Our flat MoE baseline is closely related to the approach in \citep{gururangan2023scaling}, except we used a discriminative router as opposed to $k$-Means, and we used the first 32 tokens of a document (and the features from the base Transformer LM) as the input to the router, as opposed to SVD of tf-idf vectors.

Finally, the idea to iteratively shard the data by domain to train models in parallel, followed by a model merging phase has also been proposed by~\citet{li2022branchtrainmerge}. In this work, we take this approach a step further by learning a mixture of experts model, and by automatically (and even discriminatively) sharding the data.

\section{Limitations}

The most salient limitation to \Model{} is with respect to FLOP efficiency.  In this work we made no effort to optimize its FLOP efficiency; and in the results presented, \Model{} is significantly less FLOP efficient per evaluation perplexity than a standard dense compute optimal model.   

This work also does not study scaling laws of \Model{}.  We work at a single path-size scale, and on one dataset.  

We also made no effort in this work to optimize deployment of \Model{}.  In particular, if we were to naively route more frequently during deployment as in Section (\ref{sec:freq_gating_test}), we would need to re-compute the $KV$-cache for each query after each re-route.

The severity of these limitations depends on the constraints of the training and deployment infrastructure, as well as on the task at hand. 
For instance, they may matter less when the computing infrastructure consists of hundreds of small and poorly connected islands of devices for tasks where sequences can be mostly associated to one (or very few) domains.
In general, addressing these limitations constitutes avenue of future research.

\section{Conclusions and future work} \label{sec:conclusions}

In this work, we architect an ML model as a union of modules, and we define input-output mappings that are paths through these modules.  We show that we can train paths (almost) independently, and (virtually) recombine these paths back into the desired large model.
In particular, we were able to show that distributed training by path can match the performance of a 1B dense language model for a similar wall clock time.

The design of the algorithm and architecture is guided by practical trade-offs. In our work we assume that communication is costly but compute is not. In other words, the compute that counts the most is the one of co-located devices that host a path. As long as communication between such relatively small islands of devices is limited, we can scale by distributing computation across a large number of such islands.

We introduced two key ingredients to operate under these constraints: 1) we route coarsely at the sequence level and pre-shard data by path, and 2) we extend DiLoCo to update  parameters of modules shared by multiple paths.  We believe that by combining these two ideas we have made a step towards a potentially much more scalable paradigm to train large-models.

There are several avenues of future work.  In our investigations we made no effort to control the FLOP efficiency of \Model{}, focusing on wall-clock efficiency.  Naturally, there are several straightforward design choices that could increase FLOP efficiency, for example allowing some paths to co-locate. In addition, our approaches to sharding are fairly simple, and there are many interesting possibilities for more sophisticated sharding.  \Model{} was designed with continual learning in mind, and we would like to apply it in that setting eventually.  Last but certainly not least, scaling up is of utmost interest.
Shared with ~\cite{pathways, raffel_acm23}, our long-term dream is to further refine this approach and produce a never-ending, community-driven, modular learning system that can be used by everyone to compose new predictors out of existing modules, and thus efficiently develop entirely new models and capabilities in a positive feedback loop.

\bibliographystyle{abbrvnat}
\nobibliography*
\bibliography{template_refs}

%

\section*{Acknowledgements}
We would like to thank Owen He, Diego de las Casas, Ross Hemsley, Elena Gribovskaya, Yee Whye Teh, Yutian Chen, Alexandre Galashov, Bogdan Damoc, Shreya Pathak, Amal Rannen-Triki, Alek Andreev, David Budden, Jörg Bornschein for their valuable feedback throughout the development of the project.
We would also like to thank Jeff Dean, Satinder Baveja, and Raia Hadsell for their suggestions, feedback and support. Finally, we would like to thank Kitty Stacpoole and Guy Scully for their program management support.
 
\newpage
\section*{Supplementary Materials} \label{sec:supp}

\subsection{Additional Details} \label{sec:supp_details}

\begin{table}[t]
\centering
\caption{\textbf{Hyperparameter for Inner Optimization}.}
\label{tab:inner_optim_hp}
\resizebox{1.0\linewidth}{!}{%
\begin{tabular}{@{}l|cc@{}}
\toprule
Hyperparameter &  150M model & 1B model \\
\midrule
\# Layers & 12 & 24 \\
Hidden dimension & 896 & 2048 \\
\# Heads & 16 & 16 \\
Key/value size & 64 & 128 \\
Weight decay & 0.1 & 0.1 \\
Max learning rate & $4e^{-4}$ & $2e^{-4}$ \\
Batch size & 512 & 512\\
\# Warmup steps & 1000 & 1000\\
\bottomrule
\end{tabular}
}
\end{table}


We list in \autoref{tab:inner_optim_hp} the hyperparameters used in the inner optimization. The \textit{150M model} hyperparameters are used for both the dense baselines of that size and for the path optimization of \Model. In all our experiments, we used as outer optimizer  Nesterov~\citep{sutskever2013nesterov} with outer learning rate of $0.7$ and outer momentum of $0.9$, following the same recipe introduced  by~\citet{douillard2023diloco}.

\subsection{Routing Details} \label{app:routing}

We first discuss the details of discriminative coarse routing, as used in training.  We then discuss the details of more frequent routing, as optionally used in evaluation.

\subsubsection{Document level discriminative routing}
\label{sec:disc_routing_details}
For discriminative routing, we assume that we have 
\begin{itemize}
    \item Trained an initial language model that can be used as an initialization for finetuning experts.
    \item Set up a feature extractor that gets a feature for each document to feed to the router.
\end{itemize}
We alternate between two stages:
\begin{enumerate}
    \item Train a gater, perhaps based on the PPL of current experts on an unused portion of the training set,
    \item Train experts based on a sharding from current gater.
\end{enumerate}

In this work, the feature for the router is always the average of the hidden state from the last transformer block from the initial LM over the first 32 tokens of a document; let us denote this as $g(\text{document})$.  We also reserve a small subset of documents (from here on called the ``router data'') for making decisions about router quality and training the router; we use .005 of C4 for router data.

Our initial router is always constructed via k-means; and we train the first set of paths using that router (and sharding by argmin to cluster centroid).  

To train the discriminative router, we first compute the summed auto-regressive log-likelihood of each document in the router data if encoded by each path.  That is, given $n$ documents of length $L$ and paths $f_1,..., f_K$ we get an $n \times L \times K$ array of scores $S_{ijp}$ where 
\[S_{ijp} = \log p_{f_i}(t_{l,j}| t_{l-1, j}, ..., t_{1, j}), \]
and where $t_{i, j}$ is the $j$th token in the $i$th document. 
The router is always trained using a $K$ class linear logistic classifier with \[\text{argmax}_p \sum_{j=1}^L S_{ijp}\] as the target and $g(\text{document}_i)$ as the feature.  

With large numbers of paths, we saw that the paths that were assigned the fewest documents via $\text{argmax}_p \sum_j S_{ijp}$ were even more under-represented by the output of the trained logistic regressor, and it would often happen that a path was empty when using the output of the regressor.   To remedy this, we trained a bias term to match the target document-to-path distribution.

We find that 
\begin{itemize}
    \item One step of training the discriminative router leads to a significant improvement in validation perplexity, more alternating steps lead to further minor improvements.
    \item Gains are larger with more paths.
\end{itemize}

\begin{figure*}[h!]
\centering
\includegraphics[width=.97\textwidth]{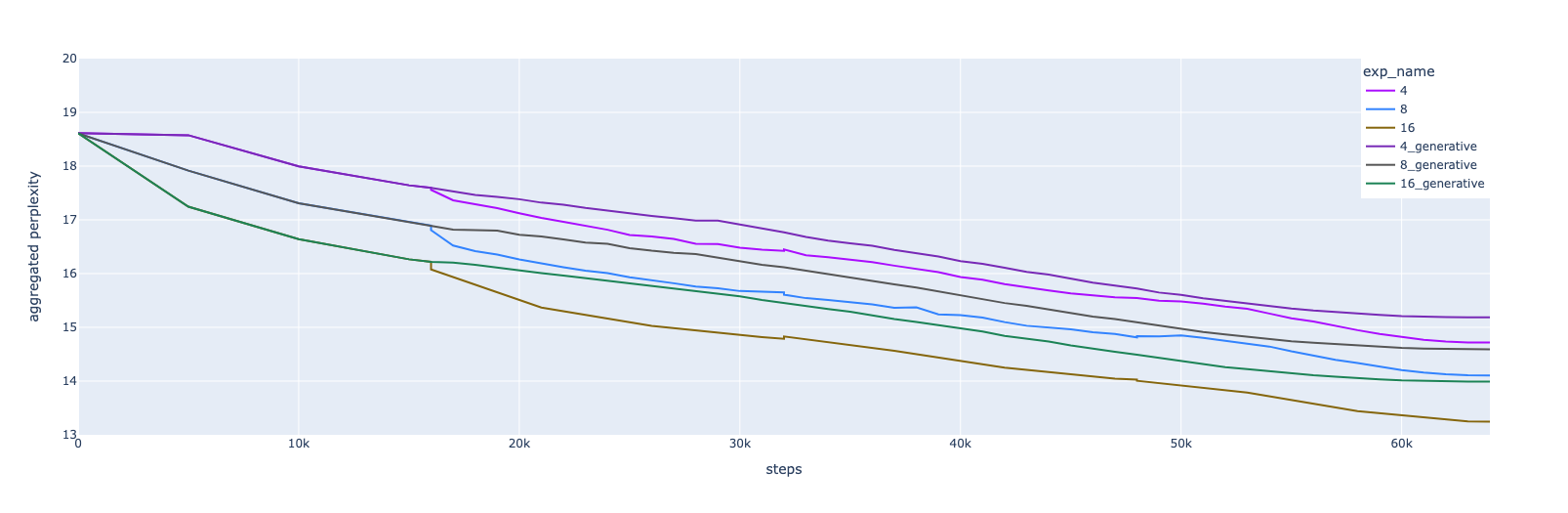}
\caption{Generative and discriminative flat MoE models with varying number of paths, three alternating discriminative phases.  The ``branching'' structure of the figure is due to the ancestry of the models: at the far left, all share an initial dense ancestor.  At 16K steps, each discriminative-generative pair of experiments shares a generative ancestor, this is continued (restarting the cosine learning rate decay) with the same router (``generative'') or with a sequence of new routers (``discriminative'')}
\label{fig:num_experts}
\end{figure*}

\begin{figure*}[h!]
\centering
\includegraphics[width=.97\textwidth]{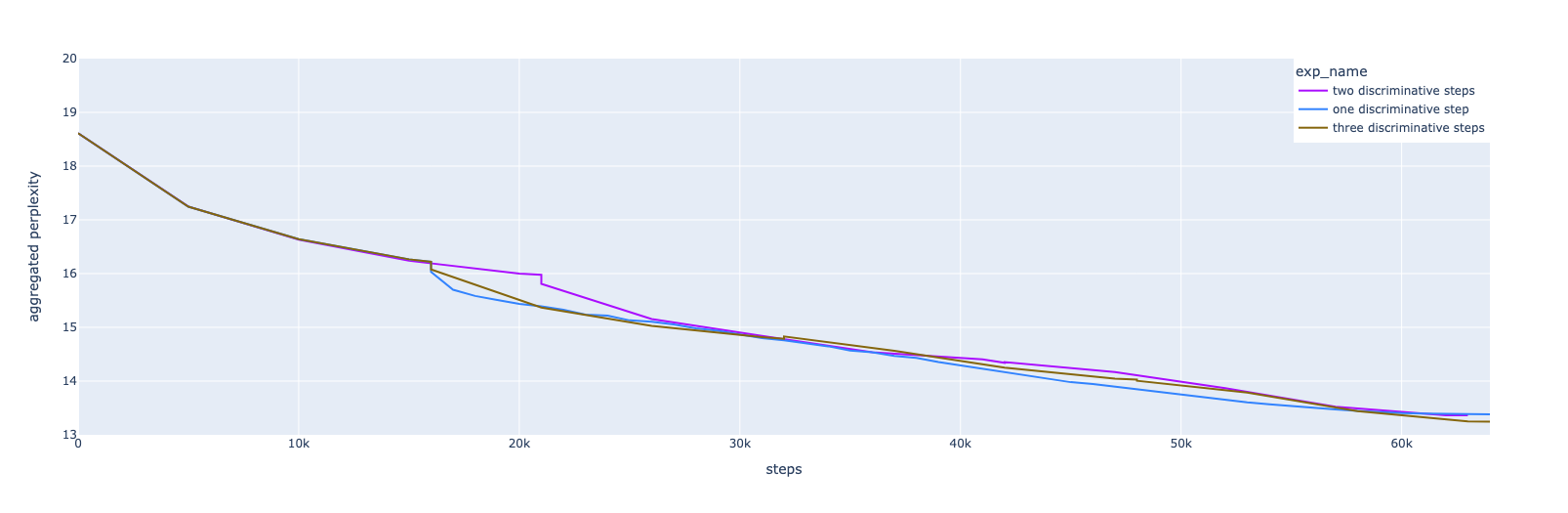}
\caption{Validation perplexity vs number of alternating minimization steps, 16 path flat MoE. As the number of alternating steps increases, the results improve, but each step gives less improvement:  $14.0 \rightarrow 13.38 \rightarrow 13.36 \rightarrow 13.25$ PPL for phases 0, 1, 2, 3 respectively.  All discriminative routing results in the main text use one alternating minimization phase.}
\label{fig:num_steps}
\end{figure*}

\subsubsection{More frequent routing during evaluation}
As before, we have the array $S_{ijp}$ giving the scores of each token in the router data set of documents.  We fix a number $L$, and build a sequence transduction training set where the inputs are the token sequences $t_{i,j}$ from the reserved portion of the training documents, and the outputs $T=T_{ij}$ are 
defined by
\[T_{ij} = \text{argmax}_p \sum_j^{j'} S_{ijp},\]
where $j' =  \min(\text{length of document}, j+L-1)$, and $L$ is a window size.   Once we have $T$, we can train a transformer to transduce $T$ from $t$.  We finetune this router from the same model that we fork to train the paths, using the same hyperparameters as our standard LM training (except no warmup), and it converges after a 2K steps.  The learned router can then make a decision at any token.

We found that choosing $L$ to be the same size as the number of tokens before re-routing (as in Table \ref{tab:frequent_gating_moeml}) led to the best results, but was only marginally better than choosing $L$ large enough so the window always went to the end of the document.  The results in Table \ref{tab:frequent_gating_moeml} use this choice of $L$ (the length of the whole sequence, $1024$).  We then can use the same learned router for any choice of evaluation-time frequency.

In \autoref{tab:sharding_moeml} we show the effect of different {\em sharding methods} on a $8 \times 8$ \Model. We observe that sharding makes a significant difference, with the discriminative variant yielding the best generalization, with an absolute gain of $0.7$ perplexity points over the generative $k$-means variant. We found that longer training runs yield even greater gain from the discriminative gating.

\subsection{More sophisticated generative routing}
\begin{table}[t]
\centering
\caption{\textbf{Sharding Impact on 8x8 \Model{}}: Validation perplexity after 32 outer optimization steps, each consisting of 62 inner optimization steps with a 8x8 \Model{} with $P=64$. Longer training further increases the gap between methods. The discriminative method is based on product $k$-Means.}
\label{tab:sharding_moeml}
\begin{tabular}{@{}l|c@{}}
\toprule
Sharding & Validation Perplexity \\
\midrule
$k$-Means & 17.2 \\
Product $k$-Means & 16.8 \\
Discriminative & 16.5 \\
\bottomrule
\end{tabular}
\end{table}

We explored Product $k$-Means as generative router for \Model{}.
When $k$ is large, $k$-Means can become inefficient, because some clusters are assigned few examples.  Moreover, in our use case, the structure of the shards does not match the structure of the modules.
Product $k$-Means is a way to mitigate these issues.  For a 2 level \Model{}, the feature $z \in \mathcal{R}^D$ is divided in two groups: $z = [z_1, z_2]$, with $z_1$ comprising the first $D/2$ dimensions, and $z_2$ the second set of $D/2$ dimensions.\footnote{We could create more groups, but we only used two in this work.} We then perform $k$-means on each set of features independently, and assign each sequence $z$ to the pair $(i, j)$. The first index refers to the assignment of $z_1$, and the second index to the assignment of $z_2$ as per equation~\ref{eq:kmeans}. Although each assignment takes only $k$ possible values, the number of possible unique pair-assignments is $k^2$. Moreover, the cost of assignment grows with the square root of the number of total pair assignments. We used this approach when experimenting with hundreds of paths; while the results were an improvement over simple $k$-Means, discriminative routing was better still; and using the product $k$-means as a router for the initial module training before the discriminative step did not significantly improve results.  Note that the discriminative routing adapts to the structure of the \Model{}, because the paths are used to score documents to find the targets.  Nevertheless, we consider alternative generative routing approaches important for further study, as discriminative scaling by scoring all paths cannot scale to large numbers of paths.

\end{document}